\def\ARXIV{}  
\documentclass[10pt,twocolumn,letterpaper]{article}

\ifdefined\ARXIV
  \usepackage[pagenumbers]{cvpr}
\else
  \usepackage[review]{cvpr}
\fi

\usepackage{cuted} 

\usepackage{currfile} 

\usepackage{caption} 

\newcommand{\nTasks}{7}
\newcommand{\tfMeanTF}{0.719}
\newcommand{\astraMean}{0.711}
\newcommand{\tfChest}{0.552}
\newcommand{\tfPneu}{0.822}
\newcommand{\tfBreast}{0.670}
\newcommand{\tfOrganMin}{0.731}
\newcommand{\tfOrganMax}{0.782}

\newcommand{\tfUterus}{0.764}
\newcommand{\tfUterusAcc}{0.698}
\newcommand{\tfRepUterus}{0.764$\pm$0.073}
\newcommand{\astraUterus}{0.136}
\newcommand{\adamMean}{0.828}
\newcommand{\gapAdam}{10.9}
\newcommand{\astraGapAdam}{11.7}
\newcommand{\gapAdamRangeBoth}{6--20}
\newcommand{\tfBeatsVLM}{5}
\newcommand{\tfVLMn}{6}
\newcommand{\vlmZSPneu}{0.273}
\newcommand{\utVZS}{0.818}
\newcommand{\utAdam}{0.996}
\newcommand{\archMin}{591}
\newcommand{\archMax}{6{,}974}
\newcommand{\utN}{293}
\newcommand{\utTrain}{117}
\newcommand{\utVal}{59}
\newcommand{\utTest}{117}
\newcommand{\utNante}{146}
\newcommand{\utNretro}{147}
\newcommand{\utDupRemoved}{12}
\newcommand{\utNvalid}{305}
\newcommand{\agClsPct}{99.1}
\newcommand{\agClsK}{0.94}
\newcommand{\wZeroMin}{22}
\newcommand{\wZeroMax}{94}
\newcommand{\wDistMin}{8}
\newcommand{\wDistMax}{145}
\newcommand{\aDistMin}{554}
\newcommand{\aDistMax}{2{,}865}
\newcommand{\fgPneuOp}{$+$0.41}
\newcommand{\fgPneuAs}{$+$0.65}
\newcommand{\fgBreastOp}{$+$0.73}
\newcommand{\fgBreastAs}{$-$0.04}
\newcommand{\fgKidOp}{$+$0.31}
\newcommand{\fgKidAs}{$+$0.43}
\newcommand{\fgLivOp}{$-$0.02}
\newcommand{\fgLivAs}{$-$0.37}
\newcommand{\fgUtOp}{$+$0.34}
\newcommand{\fgUtAs}{$-$0.48}
\newcommand{\probeAsLow}{0.121--0.241}
\newcommand{\probeClLow}{0.769--0.923}
\newcommand{\brPrior}{73.1}
\newcommand{\pnPrior}{62.5}
\newcommand{\chPrevMin}{0.2}
\newcommand{\chPrevMax}{17.6}
\newcommand{\brBiasOp}{+1.2--2.0}
\newcommand{\brBiasAs}{+0.8--1.0}
\newcommand{\exAnsAUC}{0.668}
\newcommand{\outTokMax}{79k}

\newcommand{\repourl}{\ifdefined\ARXIV\url{https://github.com/MaximilianLindholz/weight-space-programming-medical}\else (link withheld for review)\fi}

\definecolor{cvprblue}{rgb}{0.21,0.49,0.74}
\usepackage[pagebackref,breaklinks,colorlinks,allcolors=cvprblue]{hyperref}

\def\paperID{*****}
\def\confName{CVPR}
\def\confYear{2027}

\title{Weight-Space Programming for Medical Imaging:\\Do We Still Need Training Data?}

\ifdefined\ARXIV
\author{%
Maximilian Lindholz$^{1,2,*}$\hspace{0.8em}
Richard Ruppel$^{1,*}$\hspace{0.8em}
Annette Milnik$^{3}$\hspace{0.8em}
Sebastian Halskov$^{1}$\\
Semil Eminovic$^{1}$\hspace{0.8em}
Charlie A.\ Hamm$^{1,2}$\hspace{0.8em}
Robin Schmidt$^{1,4}$\hspace{0.8em}
Anika Knupfer$^{5,6}$\\
Jana Hutter$^{5,6,7}$\hspace{0.8em}
Johanna P.\ M\"uller$^{7}$\hspace{0.8em}
Carolina Dominguez Aleixo$^{1}$\hspace{0.8em}
Jawed Nawabi$^{1}$\\
Jens Vogel-Claussen$^{1,5,8}$\hspace{0.8em}
Tobias Penzkofer$^{1,2}$\\[6pt]
{\small $^1$Charit\'e -- Universit\"atsmedizin Berlin, Germany}\\
{\small $^2$Berlin Institute of Health at Charit\'e -- Universit\"atsmedizin Berlin, Germany}\\
{\small $^3$University of Basel, Switzerland}\\
{\small $^4$Experimental and Clinical Research Center (ECRC) of Charit\'e and Max Delbr\"uck Center, Berlin, Germany}\\
{\small $^5$Hannover Medical School, Germany}\\
{\small $^6$Leibniz University Hannover, Germany}\\
{\small $^7$King's College London, United Kingdom}\\
{\small $^8$German Center for Lung Research (DZL), Hannover, Germany}\\[4pt]
{\small $^*$Equal contribution.\quad Correspondence: {\tt maximilian.lindholz@charite.de}}
}
\else
\author{Anonymous CVPR submission\\Paper ID \paperID}
\fi

\begin{document}
\maketitle
\begin{abstract}
Can a large language model (LLM) write the parameters of a medical image classifier from medical knowledge alone, without any training data? Such classifiers normally require labelled images, which are costly and often unavailable. We investigate this question through weight-space programming (WSP), in which an LLM writes or adapts the weights and biases of a neural network directly, without optimisation on training data. Claude Opus~5.5 and GPT-6 Astra each generate the parameters of a small LeNet-style CNN in a single reply, without tool use and without access to any image. We compare these networks against zero-shot MedSigLIP and against the same architecture trained on labels using the Adam optimiser, across \nTasks{} classification tasks in four modalities: chest X-ray, breast ultrasound and abdominal CT (all from MedMNIST), and pelvic MRI (SynthUterus). The LLM-written networks perform above chance on all six MedMNIST tasks, with a mean test AUC of \tfMeanTF{} for Opus and \astraMean{} for Astra. Both models outperform MedSigLIP on \tfBeatsVLM{} of \tfVLMn{} tasks while using more than $10^{4}$ times fewer parameters. However, the same network trained with Adam on only 256 labelled images achieves a higher AUC on every MedMNIST task (mean AUC \adamMean{}). On SynthUterus, Opus attains an AUC of \tfUterus{}, whereas Astra falls below chance (AUC \astraUterus{}) owing to a systematic error. These results show that LLMs can write functional medical classifiers, but that they do not yet replace training data. We release the protocol, logs and all generated weights at \repourl.
\end{abstract}
\section{Introduction}
\label{sec:intro}

\begin{figure*}[t]
  \centering
  \includegraphics[width=\textwidth]{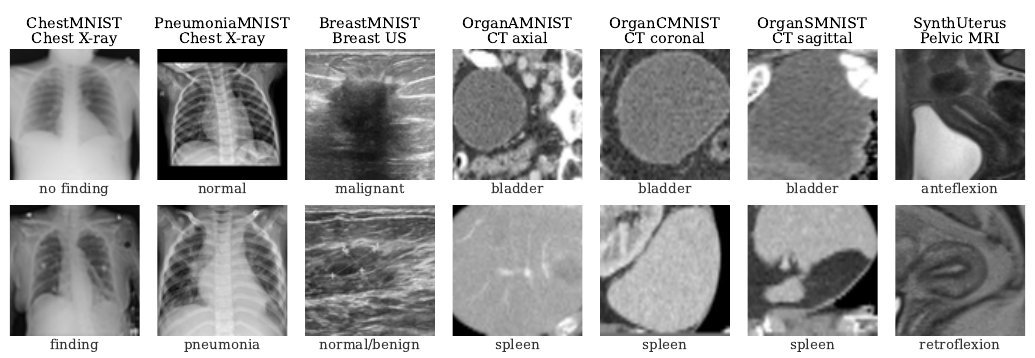}
  \caption{The \nTasks{} medical imaging tasks (\cref{sec:tasks}), two examples each. For display only, images are shown at higher resolution (MedMNIST+~\cite{yang2023medmnist} at 64 or 224 pixels,
SynthUterus at its released 96); the networks see the same images at $28{\times}28$.}
  \label{fig:tasks}
\end{figure*}

Neural networks are usually programs that are not written but
found by an optimiser on labelled data~\cite{lecun2015deep}. Takehi \etal recently introduced \emph{weight-space programming}~\cite{wsp2026}, which asks
whether frontier LLMs can now operate in the other direction, reading a network's weights and writing new ones
directly. Setting weights without gradient descent is not new in itself: hypernetworks have generated
weights since 2016~\cite{ha2017hypernetworks,knyazev2021parameter}, but they are trained predictors. In WSP, the
predictor is a general-purpose LLM that writes weights from knowledge stated in natural language.

Medical imaging is a natural test for this idea. Labelling medical images is expensive and requires radiologists, yet the rules for reading them, the relevant ``program'', are largely written down in textbooks and the medical literature. For example, a retroflexed uterus bends backward against the cervix~\cite{ameer2025anatomy}.

If an LLM can turn such knowledge into a model, a classifier could be created before the first label is collected, and every weight could carry a stated radiological rationale. Unlike the games and other artificial environments that Takehi \etal~\cite{wsp2026} use to motivate WSP, medical imaging has no source code for its rules. The imaging process itself is reproducible, but no program specifies how anatomy and disease appear in an image, or how an image maps to a diagnosis. These rules must come from the model's medical knowledge, and they must survive the gap between textbook descriptions and pixel data.

We therefore ask three questions. First, can an LLM write a working medical imaging classifier from scratch, without any task-specific training data, and do different LLMs write similar ones? Second, do we still need data: how does a written network compare with label-free foundation models and with training on a few hundred labels? Third, which design decisions do LLMs make in the parameters, and how do they compare with what gradient descent learns in the same network?

\paragraph{Contributions.}
\begin{itemize}[nosep,leftmargin=*]
  \item To our knowledge, the first study of weight-space programming in medical imaging, covering \nTasks{} tasks in four modalities (X-ray, ultrasound, CT, MRI).
  \item Evidence that networks written by both LLMs are above chance on all six MedMNIST tasks and beat zero-shot MedSigLIP on \tfBeatsVLM{} of \tfVLMn{} tasks, but stay below the same network trained on 256 labels.
\item A weight-level comparison of LLM-written and gradient-trained networks of identical architecture, covering weight statistics, learned filters and spatial attribution. The LLM-written networks use few distinct, rounded values and textbook filters, unlike the trained networks.
\end{itemize}
\section{Related Work}
\label{sec:related}

\paragraph{Predicting weights instead of training them.}
One network setting the weights of another is an established idea, dating back to Schmidhuber's fast weights, where a slowly learning network outputs the weight changes of a second, fast network at every time step~\cite{schmidhuber1992learning}.
Later milestones differ mainly in what the weights are predicted \emph{from}. In 2016, dynamic filter networks generated convolution filters from the input
image~\cite{debrabandere2016dynamic}, and HyperNetworks, which named the idea, generated a main network's weights
from learned layer embeddings~\cite{ha2017hypernetworks}. Meta networks derived fast weights from the loss gradients
of a few labelled examples for one-shot learning~\cite{munkhdalai2017meta}. Graph hypernetworks predicted all weights
of unseen architectures from their computational graph~\cite{zhang2019graph}, and GHN-2 sets the 24M parameters of a
ResNet-50 in a single forward pass~\cite{knyazev2021parameter}. Task-conditioned hypernetworks map a learned task
embedding to weights~\cite{vonoswald2020continual}, and HyperDreamBooth predicts personalisation weights of a
text-to-image model from one photo, refined by brief fine-tuning~\cite{ruiz2024hyperdreambooth}. A review sorts
hypernetworks by whether the task, the data or noise conditions them~\cite{chauhan2024brief}. In all of these, the
mapping to weights is itself learned, from labelled data and populations of trained networks. WSP~\cite{wsp2026} is
closest to a task-conditioned hypernetwork, but the task embedding is a natural-language description, and the mapping is a general-purpose LLM that was not trained for this purpose.

\paragraph{Learning in weight space.}
A growing literature treats trained networks as data. Unterthiner \etal~\cite{unterthiner2020predicting} predict
accuracy from weight statistics, and Sch\"urholt \etal~\cite{schurholt2021self} learn representations of model zoos.
Equivariant architectures process weights while respecting neuron permutation symmetries~\cite{navon2023equivariant,
zhou2023permutation}. Weight-space \emph{editing} includes task arithmetic~\cite{ilharco2023editing}, model
soups~\cite{wortsman2022soups} and locate-and-edit methods for language models~\cite{meng2022locating}.
Like the weight predictors above, these methods learn or compute their edits from data or from other networks;
in WSP, the LLM reasons about the weights itself.

\paragraph{Weight-space programming.} Takehi \etal~\cite{wsp2026} introduced the term and reported that a frontier LLM (GPT-6 Astra) can read, repair and write network weights, ranging from MinAtar and Go policies to edits of a 0.8B-parameter language model. Concurrently, Claude, operating as a tool-using coding agent, set the weights of small transformers by hand for ten algorithmic tasks~\cite{singh2026wsp}. Both reports are informal (blog posts) and have been criticised: networks written from raw weights alone are small, and most results depend on tools and forward passes~\cite{honnibal2026misleading,goedecke2026superhuman}. We adopt the most restrictive setting of~\cite{wsp2026}, a single reply without tools, audit each call, and apply it to a domain whose rules are not available as code.

\paragraph{Hand-written and compiled networks.}
Tracr~\cite{lindner2023tracr} compiles programs into transformer weights, and classic computer vision hand-designed
filters. WSP differs in that the ``compiler'' is an LLM working from natural-language domain knowledge.

\paragraph{Label-free medical classification.}
Zero-shot vision--language models such as CLIP~\cite{radford2021clip} and BiomedCLIP~\cite{zhang2023biomedclip} also
classify without task labels, but they rely on large-scale pretraining. WSP needs no labelled images and no image–text pretraining for the network it produces, although the LLM that writes it is itself pretrained on large text corpora, and it yields a network small enough to be fully inspected. We compare against the medical vision--language model
MedSigLIP~\cite{sellergren2025medgemma}.
\section{Method}
\label{sec:method}

\subsection{Datasets}
\label{sec:tasks}

We study \nTasks{} imaging tasks from two sources. MedMNIST v2~\cite{yang2023medmnist,yang2021medmnist} is a collection of standardised 2D biomedical classification datasets at $28{\times}28$ pixels, with official splits, metrics and ResNet reference results~\cite{he2016resnet}. We use all six of its 2D radiology tasks:
\emph{ChestMNIST}, which of 14 findings are present on an adult chest X-ray (multi-label: each finding is its own yes/no output~\cite{wang2017chestx}); \emph{PneumoniaMNIST}, pneumonia \vs normal on paediatric chest X-rays~\cite{kermany2018identifying}; \emph{BreastMNIST}, malignant \vs normal or benign on breast ultrasound~\cite{aldhabyani2020dataset}; and \emph{OrganA/C/SMNIST}, which of 11 abdominal organs a CT crop shows, in axial, coronal and sagittal slices (multi-class; LiTS volumes~\cite{bilic2023lits} with organ bounding boxes from~\cite{xu2019efficient}). These datasets have been public since 2020 and are widely used, so they may be in the LLMs' training data (high exposure, $^\bullet$).

The second dataset is a predefined subset of the synthetic uterine MRI dataset SynthUterus~\cite{mueller2026synthuterus}
($96{\times}96$ sagittal crops generated by orientation-conditioned diffusion models~\cite{mueller2025diffusing}, 200 per
combination of flexion and version), used for a binary task (\cref{fig:uterus}): anteverted/anteflexed \vs
retroverted/retroflexed uterus, \ie whether the corpus bends forward or backward against the cervix. We chose it a priori because the
$28{\times}28$ format needs clear geometric differences, so mixed version/flexion subclasses are excluded, as are
near-duplicate images that this resolution cannot tell apart. Labels are derived from the independent ratings of both radiologists (class agreement \agClsPct\%, $\kappa=\agClsK$). Of the 400
images of the two unmixed classes, \utNvalid{} were rated realistic, correctly classed and anomaly-free by both readers;
removing near-duplicates (shift-tolerant pixel correlation $r>0.95$ with another image of the same class,
\utDupRemoved{} images) leaves \utN{} cases (\utNante{} anteflexed, \utNretro{} retroflexed). The data have been public
only since February 2026 and are rarely downloaded (low exposure, $^\circ$). The released crops follow the radiological
convention for sagittal slices (anterior left, superior up), which the prompt does not state; we resize them to $28{\times}28$.

All tasks are scored with MedMNIST's official metrics~\cite{yang2023medmnist}: the area under the ROC curve (AUC, averaged over findings or classes) and accuracy (ACC). Below, ``the six MedMNIST tasks'' always means the first group; SynthUterus is reported separately.

\subsection{A network small enough to write}
\label{sec:arch}
Before prompting the LLMs, we fix one base architecture that is small enough to fit into their context window. We call it \emph{ProgNet} (``programmable network''): a LeNet-style CNN~\cite{lecun1998gradient} with a single convolutional stage (\cref{tab:arch}). With one stage, every weight has a direct meaning (a filter, an image region or a class), so an LLM can write it as a radiological rule. In deeper networks, layers combine features of features, which no longer map onto such rules.

The input is a greyscale image $x\in[-1,1]^{C\times28\times28}$ with $C{=}1$; pixel values $v\in\{0,\dots,255\}$ are rescaled to $v/127.5-1$. ProgNet processes it in four steps:
\begin{enumerate}[nosep,leftmargin=*]
  \item \textbf{Convolution.} $K{=}10$ filters $w_k\in\mathbb{R}^{C\times3\times3}$ with biases $b_k$ (stride 1, zero padding 1) give feature maps $z_k = w_k * x + b_k\in\mathbb{R}^{28\times28}$.
  \item \textbf{ReLU.} $a_k=\max(0,z_k)$, so each map shows where filter $k$ responds.
  \item \textbf{Pooling.} Non-overlapping $4{\times}4$ average pooling gives $g_k=\mathrm{Pool}_4(a_k)\in\mathbb{R}^{7\times7}$. Each cell is the mean response of filter $k$ in one $4{\times}4$ block of the image.
  \item \textbf{Read-out.} Each output $o$ has one weight map $M_{o,k}\in\mathbb{R}^{7\times7}$ per filter and a bias $c_o$:
\end{enumerate}
\begin{equation}
  f(x)_o = \sum_{k=1}^{K}\sum_{i,j=1}^{7} M_{o,k}[i,j]\; g_k[i,j] \;+\; c_o .
  \label{eq:prognet}
\end{equation}
Each entry $M_{o,k}[i,j]$ states how much evidence filter $k$ at grid position $(i,j)$ provides for class $o$. This is the unit in which rules are written (\cref{sec:step1}).

The output layer depends on the task: one logit with a sigmoid for binary tasks, a softmax for the multi-class organ tasks, and 14 independent sigmoids for multi-label ChestMNIST. With $n$ outputs, ProgNet has $K(9C{+}1)+n(49K{+}1)$ learnable parameters (weights and biases, together called \emph{weights} below), \ie \archMin--\archMax{} per task. ProgNet is a standard CNN that can be trained with the Adam optimiser ~\cite{kingma2015adam}, and it is small enough for an LLM to write or read all of its weights.
\begin{table}[t]
\centering
\footnotesize
\setlength{\tabcolsep}{3pt}
\caption{ProgNet, layer by layer. $C$ input channels, $K$ filters, $n$ outputs.}
\label{tab:arch}
\begin{tabular}{@{}lllr@{}}
\toprule
Step & Operation & Output & Parameters \\
\midrule
Input & image scaled to $[-1,1]$ & $C\times28\times28$ & -- \\
1 Convolution & $K$ filters $3{\times}3$ (pad 1) & $K\times28\times28$ & $K(9C{+}1)$ \\
2 Rectification & ReLU & $K\times28\times28$ & -- \\
3 Pooling & average $4{\times}4$, stride 4 & $K\times7\times7$ & -- \\
4 Read-out & linear, $n{\times}K$ maps $7{\times}7$ & $n$ & $n(49K{+}1)$ \\
Output & sigmoid or softmax & $n$ & -- \\
\bottomrule
\end{tabular}
\end{table}

\subsection{Protocol}
\paragraph{LLM writers and prompt.}\phantomsection\label{sec:prompt}
We use two LLMs as writers. Opus~5.5 is called through Claude Code 2.1.319, Anthropic's command-line agent, in its headless (non-interactive) mode. Astra, the model used in~\cite{wsp2026}, is called through the OpenAI Responses API. Both models receive the identical prompt and the one-line system prompt ``Answer the user's request.'' For both we set the \emph{reasoning effort} to \emph{high}. This setting controls how much internal reasoning the model may do before it answers. The only other setting is an output limit of 128{,}000 tokens. Each call starts with a fresh context (no tools, no MCP servers and no stored session), so no information passes between calls or tasks. \ifdefined\ARXIV\Cref{fig:prompt} shows\else Fig.~S1 in the supplementary material shows\fi{} the complete prompt for SynthUterus. The prompts of the other six tasks differ only in the task-specific lines: dataset name and citation, MedMNIST's official dataset description quoted verbatim (source, label definition, preprocessing) instead of the flexion definition, the class names, and the output, parameter count and metric that follow from the task type (\eg 11 softmax logits and 5{,}501 parameters for OrganA, 14 independent sigmoids for Chest). Everything else, including the input encoding, the row/column convention and the edit syntax, is identical.

\subsection{Writing weights from scratch: the LLM's design decisions}
\label{sec:step1}
The LLM chooses every weight and explains its choices in its reply before it writes them. The networks it writes have two parts. The $K$ filters of step~1 are a \emph{feature bank}: the image features the network can detect. The read-out maps $M_{o,k}$ of step~4 say, for each class, which feature counts as evidence where in the image. Opus made four kinds of choices:
\begin{itemize}[nosep,leftmargin=*]
  \item \textbf{Filters.} It sets filter biases so that each filter reacts only to a range of grey values.
  \item \textbf{Read-out maps.} It compares a region with a reference, which cancels overall brightness.
  \item \textbf{Read-out biases.} It sets them from the metric and the class frequency (\cref{sec:decisions}).
  \item \textbf{Whole network.} It picks a classic method and builds it from weights.
\end{itemize}

\subsection{Compared models}
\label{sec:m-trained}
We compare the LLM-written ProgNets with two kinds of models, always in this order.
\textbf{MedSigLIP}~\cite{sellergren2025medgemma} is a medical vision--language model (${\sim}400$M-parameter image
encoder) pre-trained on medical images with their text. Used \emph{zero-shot}, it needs no labelled image of the
task: each image is assigned to the class whose short text description it matches best. 
\textbf{Adam} trains the same ProgNet via gradient descent~\cite{kingma2015adam} on 256 class-stratified labelled
images (\emph{Adam, 256 labels}) or on the full training set (\emph{Adam, full data}; capped at 20k images), with
early stopping on validation AUC. Adam uses learning rate $3{\times}10^{-3}$, weight decay $10^{-4}$ and batch size 64. 
It shows what the same network reaches with labels, and its weights are the
reference for the LLMs' decisions (\cref{sec:decisions}). Published ResNet-18 results~\cite{yang2023medmnist} are
given for orientation only.

\subsection{Comparing written and trained weights}
\label{sec:m-compare}
We compare each written network with the Adam network trained on 256 labels (SynthUterus: \utTrain{} labels, seed 0). Evidence maps use the test images (a 2{,}000-image subset for Chest and Organ).

\emph{Evidence maps.} By \cref{eq:prognet}, grid cell $(i,j)$ adds $c_o[i,j]=\sum_k M_{o,k}[i,j]\,g_k[i,j]$ to logit $o$ (for softmax outputs, relative to the mean over classes). Subtracting its mean over all other test cases from its mean over test cases of class $o$ gives a $7{\times}7$ map of where the network separates class $o$ (\cref{fig:compare}). We compare the maps of two networks by their Pearson correlation $r$ over the 49 cells.

The figures show the first of the three calls of each LLM.

\section{Experiments}
\label{sec:experiments}

We use the \nTasks{} tasks of \cref{sec:tasks}: the six MedMNIST tasks with their official splits, and SynthUterus
with a fixed split of \utTrain/\utVal/\utTest{} train/val/test images. Models and baselines follow
\cref{sec:m-trained}; neither LLM saw an image before writing.

\subsection{Written networks, label-free models and training}
\label{sec:step1res}
\begin{table*}[t]
\centering
\footnotesize
\setlength{\tabcolsep}{4.5pt}
\caption{Test AUC of LLM-written networks and baselines. \emph{LLM-written}: weights written in a single tool-free reply, mean$\pm$sd over three calls. \emph{No labels}: zero-shot MedSigLIP-448. \emph{Adam}: the same ProgNet trained on 256 labels or on the full training set; the published ResNet-18 (grey) is shown for reference. Bold: best without labels per row. $^\bullet$High, $^\circ$low risk of LLM training-data exposure. $^\ddagger$\utTrain{} labels (all training cases), mean of 3 seeds. ML: multi-label.}
\label{tab:step1}
\begin{tabular}{@{}lllcccccc@{}}
\toprule
 & & & \multicolumn{2}{c}{\textbf{LLM-written, 3 calls}} & No labels & \multicolumn{2}{c}{Adam (same network)} & \textcolor{gray}{ResNet} \\
\cmidrule(lr){4-5}\cmidrule(lr){6-6}\cmidrule(lr){7-8}
Dataset & Modality & Cls. & Opus & Astra & MedSigLIP & 256 lab. & full & \textcolor{gray}{-18} \\
\midrule
Chest$^{\bullet}$ & Chest X-ray & 14 (ML) & 0.552$\pm$0.015 & 0.538$\pm$0.008 & \textbf{0.665} & 0.553 & 0.654 & \textcolor{gray}{0.768} \\
Pneumonia$^{\bullet}$ & Chest X-ray & 2 & 0.822$\pm$0.036 & \textbf{0.854}$\pm$0.012 & 0.273 & 0.914 & 0.934 & \textcolor{gray}{0.944} \\
Breast$^{\bullet}$ & Breast US & 2 & 0.670$\pm$0.009 & \textbf{0.677}$\pm$0.067 & 0.643 & 0.820 & 0.841 & \textcolor{gray}{0.901} \\
OrganA$^{\bullet}$ & Abdominal CT & 11 & \textbf{0.782}$\pm$0.000 & 0.763$\pm$0.004 & 0.656 & 0.896 & 0.982 & \textcolor{gray}{0.997} \\
OrganC$^{\bullet}$ & Abdominal CT & 11 & \textbf{0.731}$\pm$0.017 & 0.712$\pm$0.022 & 0.673 & 0.908 & 0.983 & \textcolor{gray}{0.992} \\
OrganS$^{\bullet}$ & Abdominal CT & 11 & \textbf{0.759}$\pm$0.016 & 0.721$\pm$0.020 & 0.618 & 0.875 & 0.957 & \textcolor{gray}{0.972} \\
\midrule
\multicolumn{3}{l}{Mean (6 MedMNIST tasks)} & \textbf{0.719} & 0.711 & 0.588 & 0.828 & 0.892 & \textcolor{gray}{0.929} \\
\midrule
SynthUterus$^{\circ}$ & Pelvic MRI & 2 & 0.764$\pm$0.073 & 0.136$\pm$0.053 & \textbf{0.818} & 0.996$^\ddagger$ & = & \textcolor{gray}{--} \\
 
\bottomrule
\end{tabular}
\end{table*}

\Cref{tab:step1} gives the test AUC of all models. The networks written by both LLMs are above chance on all six
MedMNIST tasks (mean \tfMeanTF{} for Opus, \astraMean{} for Astra), with a similar profile across tasks. For Opus,
the three CT organ tasks reach \tfOrganMin--\tfOrganMax, PneumoniaMNIST \tfPneu{} and BreastMNIST \tfBreast; ChestMNIST,
the mean AUC over 14 findings, stays close to chance (\tfChest). 
Results were consistent across the three independent calls, with the largest variation on BreastMNIST for Astra (0.067).

Without labels, the written networks compete with a foundation model: on the same test images, both LLMs beat
MedSigLIP zero-shot on \tfBeatsVLM{} of the \tfVLMn{} MedMNIST tasks (all except Chest), with ${\le}7$k instead of
${\sim}400$M image-encoder parameters. On Pneumonia, MedSigLIP is even below chance (\vlmZSPneu), ranking the classes
the wrong way round. With labels, training wins: the same ProgNet trained with Adam on 256 labels is better on every
MedMNIST task (mean \adamMean, \gapAdam{} AUC points above Opus and \astraGapAdam{} above Astra).

\begin{figure}[t]
  \centering
  \includegraphics[width=\linewidth]{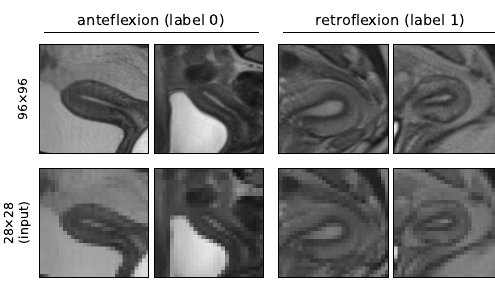}
  \caption{The SynthUterus task: anteflexion (left) \vs retroflexion (right), defined by the angle
  between the cervical canal and the uterine cavity. Two illustrative training or validation cases per class (the data owner's picks, or the
  most similar kept case where de-duplication removed a pick), at the released $96{\times}96$ resolution (top) and as the $28{\times}28$ network input (bottom). Both
  radiologists rated each case realistic, correct class, free of anomalies and artefacts.}
  \label{fig:uterus}
\end{figure}

On SynthUterus (\cref{fig:uterus}), the one task the LLMs are unlikely to have seen, Opus's networks reach
\tfRepUterus{} AUC (ACC \tfUterusAcc), below MedSigLIP zero-shot (\utVZS) and far below Adam with all \utTrain{}
training labels (\utAdam). Opus encodes the flexion as the edge orientation of cavity and canal across the whole
image. Astra's three networks are far below chance (mean \astraUterus), \ie they separate the classes the wrong way
round. Both LLMs assume the correct orientation (anterior left), and all six networks score a ``/'' line as
retroflexion and a ``\textbackslash'' line as anteflexion. Astra, however, takes the definition literally and models a
bend, comparing the orientation in the upper part of the image with that in the lower part, where it expects the
cervical canal; these crops do not show that layout, and Astra's read-out of the lower three grid rows alone gives AUC
\probeAsLow{} (Opus \probeClLow).

\subsection{LLM-written \vs Adam-trained weights}
\label{sec:decisions}
\begin{figure}[t]
  \centering
  \includegraphics[width=\linewidth]{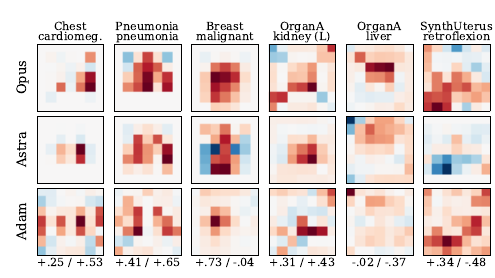}
  \caption{Where the LLM-written and Adam-trained networks (256 labels) look for selected classes (one map per class and network). Per $7{\times}7$ grid cell: contribution
  to the class logit in test cases of the class minus in the other cases (red: evidence for the class; each map scaled
  to its own maximum). Opus and Astra: first of their three calls each; below each column: correlation $r$ of
  each LLM's map with Adam's (Opus / Astra).}
  \label{fig:compare}
\end{figure}

\begin{figure}[t]
  \centering
  \includegraphics[width=\linewidth]{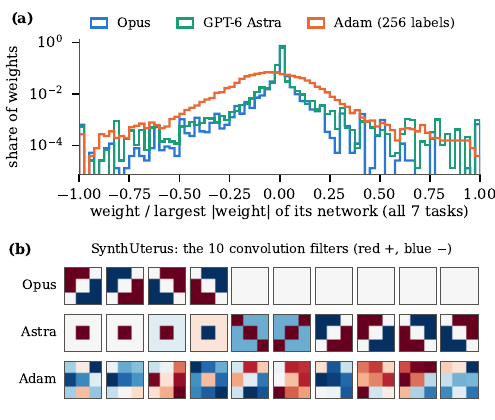}
  \caption{LLM-written \vs Adam-trained weights. (a)~Distribution of all weights of the LLM-written networks
  (Opus and Astra, three calls each) and of the 256-label Adam networks (SynthUterus: \utTrain{} labels), each scaled by its network's largest weight, pooled
  over the \nTasks{} tasks (log scale). Written weights are mostly exactly zero and take few distinct values.
  (b)~The ten SynthUterus convolution filters (first call each): Opus writes four oriented edge detectors and leaves
  six filters unused (all zero), Astra uses all ten, Adam learns smooth mixed patterns.}
  \label{fig:weights}
\end{figure}

Within the fixed architecture, each LLM decides every filter, read-out weight and bias. \Cref{fig:weights,fig:compare}
compare these decisions with the weights Adam learns for the same ProgNet from 256 labels (\cref{sec:m-compare}).

\emph{What the weights look like} (\cref{fig:weights}). Opus's networks are sparse and discrete:
\wZeroMin--\wZeroMax\% of their weights are exactly zero, and each network uses only \wDistMin--\wDistMax{} distinct
values (\eg $\pm0.25$, $\pm0.5$), whereas Adam gives every weight its own value (\aDistMin--\aDistMax{} distinct
values, none zero). Astra's networks are denser but also take few distinct values (\cref{fig:weights}a). The filters
show the same contrast (\cref{fig:weights}b): for SynthUterus, Opus writes four oriented edge detectors for the
uterine axis and leaves six filters unused, Astra fills all ten with textbook kernels (identity, centre--surround, diagonal line and edge
detectors), and Adam learns smooth, mixed patterns.

\emph{Where the networks look} (\cref{fig:compare}). Each map shows where a network separates one class from the
others. Where a finding has a typical location, the written networks often look where the trained one does: for pneumonia,
both LLMs' maps correlate with Adam's ($r=$\fgPneuOp{} for Opus, \fgPneuAs{} for Astra), as Opus's does for a
malignant breast lesion (\fgBreastOp; Astra \fgBreastAs). On CT, the agreement is weaker (left kidney \fgKidOp/\fgKidAs,
liver \fgLivOp/\fgLivAs): Opus chose coarse region features and reads laterality only from diagonal corners, whereas
gradient descent uses where an organ lies. On SynthUterus, Opus's map partly matches Adam's (\fgUtOp), while Astra's is
inverted (\fgUtAs), the same sign flip as its AUC in \cref{tab:step1}. The written maps are also sparser: the LLMs state
a few regions, whereas training spreads evidence over the whole image.

\emph{What the LLMs know about the data.} The LLMs also use what they know about each dataset's label
distribution, although no prompt states it. Opus opens its BreastMNIST answer (\ifdefined\ARXIV\cref{fig:answer}\else Fig.~S2 in the supplementary material\fi) with:
``\textbf{Prior.} About 73\% of BreastMNIST images are label 1 (normal or benign). The bias therefore starts positive,
and the network predicts `benign' unless there is clear malignant evidence.'' The true share is \brPrior\%. All six
BreastMNIST networks start from a positive output bias (Opus \brBiasOp, Astra \brBiasAs), Opus quotes the share of
pneumonia in the Pneumonia test set (about 62\%; true \pnPrior\%), and every ChestMNIST bias of both LLMs is negative,
because each of the 14 findings is rare (\chPrevMin--\chPrevMax\% of test images). On SynthUterus, such knowledge was not present: all six networks set the output bias to
zero, and Astra states that a zero bias ``avoids imposing an unsupported class-frequency prior''. An output bias
moves the decision threshold, and thus the ACC, but leaves the AUC of a binary output unchanged.

\section{Discussion}
\label{sec:discussion}

Without a single image, two LLMs turned radiological knowledge (organ density, opacified lungs, hypoechoic masses,
the axis of a flexed uterus)~\cite{singhal2023clinical} into networks that are above chance on every MedMNIST task
and beat a 400M-parameter vision--language model zero-shot on \tfBeatsVLM{} of \tfVLMn, a model class that can reach expert
level on chest X-rays at higher resolution~\cite{tiu2022expert}. They work best where the label is a matter
of anatomy and density, and worst where findings are subtle and rare, as in the 14-label chest X-rays. On SynthUterus,
the one task neither LLM is likely to have seen, MedSigLIP zero-shot does better than both. An LLM can therefore write
a working medical imaging classifier, but only in a limited sense.

The written networks also differ from trained ones in a consistent way: textbook operators, few round values and a
few stated regions. Where a finding has a typical location, the LLMs place the evidence where training does; where it
does not, they follow their own design. These choices are not one model's habit: both LLMs make the same kinds of
choices and reach a similar AUC, which suggests a shared textbook prior that, more than the writer, differs from what
data teach. The two LLMs differ in economy (Astra uses all filters) and in how
literally they take a definition: on SynthUterus, both assume the correct orientation, but Astra expects the bend at the cervix in a place where these crops do not show it, and its networks fall below chance. What the prompt leaves unstated about the images, the LLM has to guess. What it knows about a dataset, it uses: on the MedMNIST tasks, both LLMs set the output biases
from class shares that no prompt states but that they quote correctly (\cref{sec:decisions}), whereas on SynthUterus
this knowledge was not present and every bias is zero. This also strengthens our claim that SynthUterus was likely not part of the LLMs’ training data. Setting a bias from the class prior is the classic correction
for known class frequencies~\cite{saerens2002adjusting}. The biases do not change the AUC~\cite{fawcett2006introduction}, but they show that the LLMs bring knowledge of the specific datasets, not only of radiology. How much of the MedMNIST performance rests on such knowledge we cannot separate, a general problem when evaluating LLMs on public benchmarks ~\cite{sainz2023nlp}. This is why SynthUterus matters: it is unlikely to be in the models’ training data.

Training data are still needed. A few hundred labels beat the written networks on every task (mean over three calls), by
\gapAdamRangeBoth{} AUC points on the MedMNIST tasks except the 14-label chest findings, where both are near chance,
and by far more on SynthUterus (\utAdam{} \vs~\tfUterus). This is far from the results for games, where the generating
process is given as code~\cite{wsp2026}; trained weight predictors show a similar gap (GHN-2: 58.6\% \vs 93.5\% with
SGD on CIFAR-10~\cite{knyazev2021parameter}). In medical imaging, no code defines the classes, so the writer must guess
how a sign looks at $28{\times}28$ pixels and where the dataset draws the line, and no labelled image tells it whether
the guess is right. Labels thus remain the bottleneck, as curation and expert annotation dominate the cost of a
dataset~\cite{willemink2020preparing}.

What LLM-written networks offer instead is auditability. For high-stakes decisions, models should be interpretable by
construction rather than explained after the fact~\cite{rudin2019stop,chen2019looks}, and a written network is such a model: every
weight carries a stated region, feature and sign that a radiologist can review, and every prediction can be traced
cell by cell from filter response over the pooled grid to the logit. Its rules can be wrong, but they
cannot silently rely on hospital, scanner or text markers, shortcuts that learned chest X-ray classifiers
exploit~\cite{zech2018variable,degrave2021shortcuts,geirhos2020shortcut}; for hypernetworks, explaining generated weights is still
open~\cite{chauhan2024brief}. Predicted weights also help as an initialisation~\cite{knyazev2021parameter,ruiz2024hyperdreambooth};
whether written ones do, we have not tested. Written networks are thus a transparent, label-free starting point and a
reference for a trained network's decisions.

Writing weights matters most where our benchmarks cannot follow, rare diseases with too few cases to train on, which together affect
3.5--5.9\% of the population~\cite{nguengang2020estimating}, and
new diseases with only few known cases. A case report often describes what a finding looks like and where it lies before a significant amount of data exist. A written network could turn that description into a reviewable classifier on the day it is published, and rank incoming cases for expert labelling, as active learning does with trained models~\cite{budd2021survey}. 

\label{sec:limits}
Our study has limitations. The scale is small by design: $28{\times}28$ images and networks of ${\le}7$k weights,
because every weight must fit into one reply (128k output tokens including reasoning; answers take up to \outTokMax).
The models are therefore far below the state of the art, and MedSigLIP is handicapped by upsampled inputs and untuned
prompts. SynthUterus is recent, so LLM exposure is unlikely but not excluded; its images are
synthetic and few (\utN), two readers rated them without adjudication, and flexion read from a 2D slice carries some
label noise. Finally,
we test two LLMs with one prompt: other LLMs or prompts may write differently, our prompt did not describe the image
layout, and a wrong guess about it inverted Astra's SynthUterus networks. Both writers know the MedMNIST datasets (\cref{sec:decisions});
such contamination can be probed~\cite{golchin2024timetravel} but not excluded.

\paragraph{Conclusion.}
LLMs can write medical imaging classifiers without data that compete with a foundation model's zero-shot
predictions on most tasks, but a few hundred labels still beat them on average on every task.

\paragraph{Ethics.}
We use only public, de-identified data under their sources' licences: MedMNIST~\cite{yang2021medmnist,yang2023medmnist}
and SynthUterus (CC BY 4.0; synthetic images~\cite{mueller2025diffusing,mueller2026synthuterus}). No patient data were
collected.

\paragraph{Use of AI.}
Claude Opus~5.5 and GPT-6 Astra are the objects of study; Claude also helped draft text and code; the
authors take full responsibility.

\ifdefined\ARXIV
\paragraph{Funding.}
M.L., C.H., J.N.\ and T.P.\ are fellows or alumni of the BIH Charit\'e (Junior Digital/Advanced) Clinician Scientist
Program funded by Charit\'e -- Universit\"atsmedizin Berlin, the Berlin Institute of Health, and Stiftung Charit\'e.

\paragraph{Competing interests.}
T.P.\ reported receiving grant funding from the Berlin Institute of Health (Advanced Clinician Scientist Grant,
Platform Grant), Ministry of Education and Research (01KX2021 [RACOON], 01KX2121 [NUM 2.0, RACOON], 01KX2524 [NUM 3.0],
68GX21001A, 01ZZ2315D), German Research Foundation (SFB 1240/2), and European Union (H2020, CHAIMELEON: 952172;
DIGITAL, EUCAIM: 10110063) during the conduct of the study; research agreements (no personal payments) with AGO
Research GmbH, Aravive Inc, ARCAGY-GINECO, Astellas Pharma Global Development Inc, AstraZeneca AB, AstraZeneca GmbH,
Clovis Oncology, EQRx International Inc, F.\ Hoffmann-La Roche Ltd, GlaxoSmithKline Research and Development Limited,
Grupo Espa\~nol de Investigaci\'on en C\'ancer de Ovario, ImmunoGen Inc, Incyte Corporation, Karyopharm Therapeutics,
Mario Negri Gynecology Oncology Group, Merck KGaA, Merck Sharp \& Dohme, NOGGO eV, Nordic Society of Gynaecological
Oncology--Clinical Trial Unit, Novartis Pharma GmbH, NovoCure GmbH, Sutro Biopharma Inc, TESARO Inc, TORL
Biotherapeutics LLC, Tubulis GmbH, Universit\"atspoliklinik A.\ Gemelli, and Verastem Inc outside the submitted work;
and personal fees from Elsevier and Bayer Healthcare. C.H.\ reported receiving personal fees from Lucida Medical,
Bayer AG, Siemens Healthineers, Koelis GmbH, and Bracco Imaging Deutschland and grants from the Ministry of Education
and Research (01KX2524 [NUM 3.0], 01KX2626 [NUM SVIK]) outside the submitted work. J.N.\ reports research collaboration agreements with
Briya Labs Ltd., Cercare Medical, and mediaire, and has received grant funding from the European Union (101079894).
He is participant in the Berlin Advanced Clinician Scientist Program proHealth -- F\"orderung von Advanced Clinician
Scientists im Bereich Pr\"avention funded by the Federal Ministry of Research, Technology and Space, project number:
01EO2508. The other authors declare no competing interests.
\fi

{
    \small
    \bibliographystyle{ieeenat_fullname}
    \bibliography{main}

@article{lecun2015deep,
  title={Deep learning},
  author={LeCun, Yann and Bengio, Yoshua and Hinton, Geoffrey},
  journal={Nature},
  volume={521},
  number={7553},
  pages={436--444},
  year={2015},
  publisher={Nature Publishing Group UK, London}
}

@incollection{ameer2025anatomy,
  title={Anatomy, abdomen and pelvis: uterus},
  author={Ameer, Muhammad Atif and Peterson, Diana C},
  booktitle={StatPearls [Internet]},
  year={2025},
  publisher={StatPearls Publishing}
}

@misc{wsp2026,
  author       = {Takehi, Rikiya and Teoh, Jayden and Zhang, Alex and Khattab, Omar},
  title        = {Sparks of Weight-Space Programming: {LLMs} can natively read \& write neural network weights},
  howpublished = {MIT CSAIL blog post, 2 November 2026, \url{https://rikiyat.github.io/blog/weight-space-programming/}; logs at \url{https://github.com/RikiyaT/lm_understand-logs}},
  year         = {2026}
}

@misc{singh2026wsp,
  author       = {Singh, Chandan},
  title        = {Weight-Space Programming from scratch on simple tasks},
  howpublished = {Report, 2 November 2026, \url{https://csinva.io/weight-space-programming-autoresearch/analyze/toy_examples/report.html}},
  year         = {2026}
}

@misc{honnibal2026misleading,
  author       = {Honnibal, Matthew},
  title        = {The ``{Astra} can read and write weights to improve {Qwen}'' post is misleading},
  howpublished = {Blog post, 3 November 2026, \url{https://honnibal.dev/blog/astra-read-write-misleading}},
  year         = {2026}
}

@misc{goedecke2026superhuman,
  author       = {Goedecke, Sean},
  title        = {{LLMs} are not superhuman at reading and writing neural network weights},
  howpublished = {Blog post, 3 November 2026, \url{https://www.seangoedecke.com/llms-are-not-superhuman-at-reading-and-writing-neural-network-weights/}},
  year         = {2026}
}

@article{yang2023medmnist,
  title   = {{MedMNIST} v2 -- A large-scale lightweight benchmark for {2D} and {3D} biomedical image classification},
  author  = {Yang, Jiancheng and Shi, Rui and Wei, Donglai and Liu, Zequan and Zhao, Lin and Ke, Bilian and Pfister, Hanspeter and Ni, Bingbing},
  journal = {Scientific Data},
  volume  = {10},
  number  = {1},
  pages   = {41},
  year    = {2023}
}

@inproceedings{yang2021medmnist,
  title     = {{MedMNIST} Classification Decathlon: A Lightweight {AutoML} Benchmark for Medical Image Analysis},
  author    = {Yang, Jiancheng and Shi, Rui and Ni, Bingbing},
  booktitle = {IEEE International Symposium on Biomedical Imaging (ISBI)},
  year      = {2021}
}

@article{unterthiner2020predicting,
  title   = {Predicting Neural Network Accuracy from Weights},
  author  = {Unterthiner, Thomas and Keysers, Daniel and Gelly, Sylvain and Bousquet, Olivier and Tolstikhin, Ilya},
  journal = {arXiv preprint arXiv:2002.11448},
  year    = {2020}
}

@inproceedings{schurholt2021self,
  title     = {Self-Supervised Representation Learning on Neural Network Weights for Model Characteristic Prediction},
  author    = {Sch{\"u}rholt, Konstantin and Kostadinov, Dimche and Borth, Damian},
  booktitle = {Advances in Neural Information Processing Systems (NeurIPS)},
  year      = {2021}
}

@inproceedings{navon2023equivariant,
  title     = {Equivariant Architectures for Learning in Deep Weight Spaces},
  author    = {Navon, Aviv and Shamsian, Aviv and Achituve, Idan and Fetaya, Ethan and Chechik, Gal and Maron, Haggai},
  booktitle = {International Conference on Machine Learning (ICML)},
  year      = {2023}
}

@inproceedings{zhou2023permutation,
  title     = {Permutation Equivariant Neural Functionals},
  author    = {Zhou, Allan and Yang, Kaien and Burns, Kaylee and Cardace, Adriano and Jiang, Yiding and Sokota, Samuel and Kolter, J. Zico and Finn, Chelsea},
  booktitle = {Advances in Neural Information Processing Systems (NeurIPS)},
  year      = {2023}
}

@inproceedings{lindner2023tracr,
  title     = {Tracr: Compiled Transformers as a Laboratory for Interpretability},
  author    = {Lindner, David and Kram{\'a}r, J{\'a}nos and Farquhar, Sebastian and Rahtz, Matthew and McGrath, Thomas and Mikulik, Vladimir},
  booktitle = {Advances in Neural Information Processing Systems (NeurIPS)},
  year      = {2023}
}

@article{zhang2023biomedclip,
  title   = {A Multimodal Biomedical Foundation Model Trained from Fifteen Million Image--Text Pairs},
  author  = {Zhang, Sheng and Xu, Yanbo and Usuyama, Naoto and others},
  journal = {NEJM AI},
  volume  = {2},
  number  = {1},
  year    = {2025},
  doi     = {10.1056/AIoa2400640},
  note    = {BiomedCLIP; preprint arXiv:2303.00915}
}

@inproceedings{radford2021clip,
  title     = {Learning Transferable Visual Models From Natural Language Supervision},
  author    = {Radford, Alec and Kim, Jong Wook and Hallacy, Chris and others},
  booktitle = {International Conference on Machine Learning (ICML)},
  year      = {2021}
}

@inproceedings{he2016resnet,
  title     = {Deep Residual Learning for Image Recognition},
  author    = {He, Kaiming and Zhang, Xiangyu and Ren, Shaoqing and Sun, Jian},
  booktitle = {IEEE Conference on Computer Vision and Pattern Recognition (CVPR)},
  year      = {2016}
}

@inproceedings{kingma2015adam,
  title     = {Adam: A Method for Stochastic Optimization},
  author    = {Kingma, Diederik P. and Ba, Jimmy},
  booktitle = {International Conference on Learning Representations (ICLR)},
  year      = {2015}
}

@inproceedings{ilharco2023editing,
  title     = {Editing Models with Task Arithmetic},
  author    = {Ilharco, Gabriel and Ribeiro, Marco Tulio and Wortsman, Mitchell and Gururangan, Suchin and Schmidt, Ludwig and Hajishirzi, Hannaneh and Farhadi, Ali},
  booktitle = {International Conference on Learning Representations (ICLR)},
  year      = {2023}
}

@inproceedings{meng2022locating,
  title     = {Locating and Editing Factual Associations in {GPT}},
  author    = {Meng, Kevin and Bau, David and Andonian, Alex and Belinkov, Yonatan},
  booktitle = {Advances in Neural Information Processing Systems (NeurIPS)},
  year      = {2022}
}

@inproceedings{wortsman2022soups,
  title     = {Model Soups: Averaging Weights of Multiple Fine-Tuned Models Improves Accuracy without Increasing Inference Time},
  author    = {Wortsman, Mitchell and Ilharco, Gabriel and Gadre, Samir Yitzhak and others},
  booktitle = {International Conference on Machine Learning (ICML)},
  year      = {2022}
}

@article{kermany2018identifying,
  title   = {Identifying Medical Diagnoses and Treatable Diseases by Image-Based Deep Learning},
  author  = {Kermany, Daniel S. and Goldbaum, Michael and Cai, Wenjia and others},
  journal = {Cell},
  volume  = {172},
  number  = {5},
  pages   = {1122--1131},
  year    = {2018}
}

@article{aldhabyani2020dataset,
  title   = {Dataset of Breast Ultrasound Images},
  author  = {Al-Dhabyani, Walid and Gomaa, Mohammed and Khaled, Hussien and Fahmy, Aly},
  journal = {Data in Brief},
  volume  = {28},
  pages   = {104863},
  year    = {2020}
}

@inproceedings{wang2017chestx,
  title     = {{ChestX-ray8}: Hospital-Scale Chest X-Ray Database and Benchmarks on Weakly-Supervised Classification and Localization of Common Thorax Diseases},
  author    = {Wang, Xiaosong and Peng, Yifan and Lu, Le and Lu, Zhiyong and Bagheri, Mohammadhadi and Summers, Ronald M.},
  booktitle = {IEEE Conference on Computer Vision and Pattern Recognition (CVPR)},
  year      = {2017}
}

@article{bilic2023lits,
  title   = {The Liver Tumor Segmentation Benchmark ({LiTS})},
  author  = {Bilic, Patrick and Christ, Patrick and others},
  journal = {Medical Image Analysis},
  volume  = {84},
  pages   = {102680},
  year    = {2023}
}

@article{sellergren2025medgemma,
  title   = {An open vision-language model for diverse medical applications},
  author  = {Sellergren, Andrew and Kazemzadeh, Sahar and Mahvar, Fereshteh and Kiraly, Atilla and Traverse, Madeleine and others},
  journal = {Nature Medicine},
  volume  = {32},
  number  = {10},
  pages   = {3695--3704},
  year    = {2026},
  doi     = {10.1038/s41591-026-04626-w},
  note    = {MedGemma and MedSigLIP; preprint arXiv:2507.05201 (``MedGemma Technical Report'')}
}

@misc{mueller2026synthuterus,
  title        = {Synthetic Uterus {MRI} Dataset with Uterine Orientation Labels},
  author       = {M{\"u}ller, Johanna P. and Knupfer, Anika and Lindholz, Maximilian and Schmidt, Robin and Kainz, Bernhard and Hutter, Jana},
  year         = {2026},
  howpublished = {Zenodo, version 2, \url{https://doi.org/10.5281/zenodo.18587285}},
  note         = {IEEE International Symposium on Biomedical Imaging (ISBI) 2026. CC BY 4.0}
}

@inproceedings{mueller2025diffusing,
  title     = {Diffusing the Blind Spot: Uterine {MRI} Synthesis with Diffusion Models},
  author    = {M{\"u}ller, Johanna P. and Knupfer, Anika and Bl{\"o}ss, Pedro and Berardi Vittur, Edoardo and Kainz, Bernhard and Hutter, Jana},
  booktitle = {Skin Image Analysis, and Computer-Aided Pelvic Imaging for Female Health},
  series    = {Lecture Notes in Computer Science},
  pages     = {93--102},
  publisher = {Springer Nature Switzerland},
  year      = {2025},
  doi       = {10.1007/978-3-032-05825-6_9}
}

@article{lecun1998gradient,
  title   = {Gradient-Based Learning Applied to Document Recognition},
  author  = {LeCun, Yann and Bottou, L{\'e}on and Bengio, Yoshua and Haffner, Patrick},
  journal = {Proceedings of the IEEE},
  volume  = {86},
  number  = {11},
  pages   = {2278--2324},
  year    = {1998},
  doi     = {10.1109/5.726791}
}

@article{rudin2019stop,
  author  = {Rudin, Cynthia},
  title   = {Stop explaining black box machine learning models for high stakes decisions and use interpretable models instead},
  journal = {Nature Machine Intelligence},
  volume  = {1}, number = {5}, pages = {206--215}, year = {2019}
}

@article{degrave2021shortcuts,
  author  = {DeGrave, Alex J. and Janizek, Joseph D. and Lee, Su-In},
  title   = {{AI} for radiographic {COVID-19} detection selects shortcuts over signal},
  journal = {Nature Machine Intelligence},
  volume  = {3}, number = {7}, pages = {610--619}, year = {2021}
}

@article{zech2018variable,
  author  = {Zech, John R. and Badgeley, Marcus A. and Liu, Manway and Costa, Anthony B. and Titano, Joseph J. and Oermann, Eric Karl},
  title   = {Variable generalization performance of a deep learning model to detect pneumonia in chest radiographs: A cross-sectional study},
  journal = {PLOS Medicine},
  volume  = {15}, number = {11}, pages = {e1002683}, year = {2018}
}

@article{singhal2023clinical,
  author  = {Singhal, Karan and Azizi, Shekoofeh and Tu, Tao and Mahdavi, S. Sara and Wei, Jason and Chung, Hyung Won and Scales, Nathan and Tanwani, Ajay and Cole-Lewis, Heather and Pfohl, Stephen and others},
  title   = {Large language models encode clinical knowledge},
  journal = {Nature},
  volume  = {620}, number = {7972}, pages = {172--180}, year = {2023}
}

@article{willemink2020preparing,
  author  = {Willemink, Martin J. and Koszek, Wojciech A. and Hardell, Cailin and Wu, Jie and Fleischmann, Dominik and Harvey, Hugh and Folio, Les R. and Summers, Ronald M. and Rubin, Daniel L. and Lungren, Matthew P.},
  title   = {Preparing Medical Imaging Data for Machine Learning},
  journal = {Radiology},
  volume  = {295}, number = {1}, pages = {4--15}, year = {2020}
}

@article{schmidhuber1992learning,
  author  = {Schmidhuber, J{\"u}rgen},
  title   = {Learning to Control Fast-Weight Memories: An Alternative to Dynamic Recurrent Networks},
  journal = {Neural Computation},
  volume  = {4}, number = {1}, pages = {131--139}, year = {1992},
  doi     = {10.1162/neco.1992.4.1.131}
}

@inproceedings{debrabandere2016dynamic,
  author    = {Jia, Xu and De Brabandere, Bert and Tuytelaars, Tinne and Van Gool, Luc},
  title     = {Dynamic Filter Networks},
  booktitle = {Advances in Neural Information Processing Systems (NeurIPS)},
  volume    = {29},
  year      = {2016}
}

@inproceedings{ha2017hypernetworks,
  author    = {Ha, David and Dai, Andrew M. and Le, Quoc V.},
  title     = {{HyperNetworks}},
  booktitle = {International Conference on Learning Representations (ICLR)},
  year      = {2017},
  url       = {https://openreview.net/forum?id=rkpACe1lx}
}

@inproceedings{munkhdalai2017meta,
  author    = {Munkhdalai, Tsendsuren and Yu, Hong},
  title     = {Meta Networks},
  booktitle = {International Conference on Machine Learning (ICML)},
  pages     = {2554--2563},
  year      = {2017}
}

@inproceedings{zhang2019graph,
  author    = {Zhang, Chris and Ren, Mengye and Urtasun, Raquel},
  title     = {Graph {HyperNetworks} for Neural Architecture Search},
  booktitle = {International Conference on Learning Representations (ICLR)},
  year      = {2019}
}

@inproceedings{vonoswald2020continual,
  author    = {von Oswald, Johannes and Henning, Christian and Grewe, Benjamin F. and Sacramento, Jo{\~a}o},
  title     = {Continual Learning with Hypernetworks},
  booktitle = {International Conference on Learning Representations (ICLR)},
  year      = {2020}
}

@inproceedings{knyazev2021parameter,
  author    = {Knyazev, Boris and Drozdzal, Michal and Taylor, Graham W. and Romero Soriano, Adriana},
  title     = {Parameter Prediction for Unseen Deep Architectures},
  booktitle = {Advances in Neural Information Processing Systems (NeurIPS)},
  year      = {2021}
}

@inproceedings{ruiz2024hyperdreambooth,
  author    = {Ruiz, Nataniel and Li, Yuanzhen and Jampani, Varun and Wei, Wei and Hou, Tingbo and Pritch, Yael and Wadhwa, Neal and Rubinstein, Michael and Aberman, Kfir},
  title     = {{HyperDreamBooth}: {HyperNetworks} for Fast Personalization of Text-to-Image Models},
  booktitle = {IEEE/CVF Conference on Computer Vision and Pattern Recognition (CVPR)},
  pages     = {6527--6536},
  year      = {2024},
  doi       = {10.1109/CVPR52733.2024.00624}
}

@article{chauhan2024brief,
  author  = {Chauhan, Vinod Kumar and Zhou, Jiandong and Lu, Ping and Molaei, Soheila and Clifton, David A.},
  title   = {A Brief Review of Hypernetworks in Deep Learning},
  journal = {Artificial Intelligence Review},
  volume  = {57}, number = {9}, pages = {250}, year = {2024},
  doi     = {10.1007/s10462-024-10862-8}
}

@inproceedings{golchin2024timetravel,
  author    = {Golchin, Shahriar and Surdeanu, Mihai},
  title     = {Time Travel in {LLMs}: Tracing Data Contamination in Large Language Models},
  booktitle = {International Conference on Learning Representations (ICLR)},
  year      = {2024}
}

@article{xu2019efficient,
  author  = {Xu, Xuanang and Zhou, Fugen and Liu, Bo and Fu, Dongshan and Bai, Xiangzhi},
  title   = {Efficient Multiple Organ Localization in {CT} Image Using {3D} Region Proposal Network},
  journal = {IEEE Transactions on Medical Imaging},
  volume  = {38}, number = {8}, pages = {1885--1898}, year = {2019}
}

@article{saerens2002adjusting,
  title   = {Adjusting the Outputs of a Classifier to New a Priori Probabilities: A Simple Procedure},
  author  = {Saerens, Marco and Latinne, Patrice and Decaestecker, Christine},
  journal = {Neural Computation},
  volume  = {14},
  number  = {1},
  pages   = {21--41},
  year    = {2002},
  doi     = {10.1162/089976602753284446}
}

@article{fawcett2006introduction,
  title   = {An introduction to {ROC} analysis},
  author  = {Fawcett, Tom},
  journal = {Pattern Recognition Letters},
  volume  = {27},
  number  = {8},
  pages   = {861--874},
  year    = {2006},
  doi     = {10.1016/j.patrec.2005.10.010}
}

@article{geirhos2020shortcut,
  title   = {Shortcut learning in deep neural networks},
  author  = {Geirhos, Robert and Jacobsen, J{\"o}rn-Henrik and Michaelis, Claudio and Zemel, Richard and Brendel, Wieland and Bethge, Matthias and Wichmann, Felix A.},
  journal = {Nature Machine Intelligence},
  volume  = {2},
  number  = {11},
  pages   = {665--673},
  year    = {2020},
  doi     = {10.1038/s42256-020-00257-z}
}

@article{tiu2022expert,
  title   = {Expert-level detection of pathologies from unannotated chest {X}-ray images via self-supervised learning},
  author  = {Tiu, Ekin and Talius, Ellie and Patel, Pujan and Langlotz, Curtis P. and Ng, Andrew Y. and Rajpurkar, Pranav},
  journal = {Nature Biomedical Engineering},
  volume  = {6},
  pages   = {1399--1406},
  year    = {2022},
  doi     = {10.1038/s41551-022-00936-9}
}

@article{nguengang2020estimating,
  title   = {Estimating cumulative point prevalence of rare diseases: analysis of the {Orphanet} database},
  author  = {Nguengang Wakap, St{\'e}phanie and Lambert, Deborah M. and Olry, Annie and Rodwell, Charlotte and Gueydan, Charlotte and Lanneau, Val{\'e}rie and Murphy, Daniel and Le Cam, Yann and Rath, Ana},
  journal = {European Journal of Human Genetics},
  volume  = {28},
  number  = {2},
  pages   = {165--173},
  year    = {2020},
  doi     = {10.1038/s41431-019-0508-0}
}

@inproceedings{sainz2023nlp,
  title     = {{NLP} Evaluation in trouble: On the Need to Measure {LLM} Data Contamination for each Benchmark},
  author    = {Sainz, Oscar and Campos, Jon Ander and Garc{\'i}a-Ferrero, Iker and Etxaniz, Julen and Lopez de Lacalle, Oier and Agirre, Eneko},
  booktitle = {Findings of the Association for Computational Linguistics: EMNLP 2023},
  pages     = {10776--10787},
  year      = {2023}
}

@article{budd2021survey,
  title   = {A survey on active learning and human-in-the-loop deep learning for medical image analysis},
  author  = {Budd, Samuel and Robinson, Emma C. and Kainz, Bernhard},
  journal = {Medical Image Analysis},
  volume  = {71},
  pages   = {102062},
  year    = {2021},
  doi     = {10.1016/j.media.2021.102062}
}

@inproceedings{chen2019looks,
  title     = {This Looks Like That: Deep Learning for Interpretable Image Recognition},
  author    = {Chen, Chaofan and Li, Oscar and Tao, Chaofan and Barnett, Alina Jade and Su, Jonathan and Rudin, Cynthia},
  booktitle = {Advances in Neural Information Processing Systems},
  volume    = {32},
  pages     = {8928--8939},
  year      = {2019}
}
}
\ifdefined\ARXIV
\clearpage
\appendix
\section*{Supplementary material}
\setcounter{figure}{0}\renewcommand{\thefigure}{S\arabic{figure}}

\paragraph{S1. The prompt.}
\Cref{fig:prompt} shows the complete prompt for SynthUterus, the only user message of each call (\ifdefined\ARXIV\cref{sec:prompt}\else Sec.~3.3 of the main paper\fi{} describes the protocol). The prompts of the other six tasks differ only in the task-specific lines (dataset, official description, classes, output and parameter count); no prompt describes what the images look like.

\begin{figure*}[p]
\centering
\begin{minipage}{\textwidth}\scriptsize
\begin{verbatim}
YOUR TASK: write, from scratch, the weights of a small convolutional network that solves this medical image classification
  task as well as possible. There is no training: no data, no loss, no gradients, no fine-tuning; you choose every number
  yourself. You have no tools: no code execution, no calculator, nothing to run, and you will not see any image from this
  dataset. This is a single response and you will get no feedback. First describe your design briefly (which image features
  each filter detects, and which evidence each output reads where), then write the edits block.

THE DATASET: SynthUterus (Mueller et al. 2026, Zenodo): synthetic, region-of-interest-cropped uterine MRI slices; only cases
  validated by two radiologists are used; images downsampled from 96 x 96 to 28 x 28. Task type: binary-class.
Official description: Synthetic sagittal uterine MRI slices (region-of-interest crops, greyscale): uterine flexion, defined
  by the angle between the cervical canal line and the uterine cavity line: anteflexion (cavity bent forward) or
  retroflexion (cavity bent backward)?
Labels:
  0: anteflexion
  1: retroflexion

THE INPUT: one image of 28 x 28 pixels with 1 channel (grey). Each pixel value p in 0..255 is given to the network as x = (p
  / 255 - 0.5) / 0.5, i.e. in [-1, 1] (-1 = black, +1 = white). Rows go from 0 (top) to 27 (bottom), columns from 0 (left)
  to 27 (right), exactly as stored in the dataset arrays.

THE OUTPUT: one logit o_0 = log-odds that the label is 1 ('retroflexion'); the score is sigmoid(o_0).

THE NETWORK (591 parameters; the same architecture as the gradient-trained baselines it is compared with):
  h   = ReLU(conv3x3(x; CW, CB))     CW: 10 x 1 x 3 x 3, CB: 10   (stride 1, zero padding 1 -> 10 x 28 x 28)
  g   = AvgPool4x4(h)                -> 10 x 7 x 7 (cell (i, j) averages pixel rows 4i..4i+3, columns 4j..4j+3)
  o_n = sum over k, i, j of M[n, k, i, j] * g[k, i, j]  +  B[n]      M: 1 x 10 x 7 x 7, B: 1
CW[k, c] is the 3 x 3 kernel of filter k on input channel c, applied as cross-correlation: kernel entry [a, b] multiplies
  the pixel at offset (a - 1, b - 1) (row, column). EVERY WEIGHT STARTS AT 0; you set the weights you need.

HOW YOUR NETWORK IS SCORED: on the official test split with MedMNIST's official metrics: AUC of sigmoid(o_0), and ACC with
  threshold 0.5. It is compared with networks of the same architecture trained by gradient descent on 256 labelled images
  and on the full training set.

HOW TO WRITE THE WEIGHTS (numbers only; nothing is run). Write a code block whose opening line is three backticks followed
  by the word edits, one statement per line ('#' starts a comment; a long list may continue over the next lines). Tensors:
  CW, CB, M, B.

```edits
CW[0, 0] = [[0, 0, 0], [0, 1, 0], [0, 0, 0]]   # filter 0, channel 0: identity kernel
CB[0] = -0.5                                    # one entry
M[2, 0, 1:6, 1:3] = 0.1                         # output 2 reads filter 0 in grid rows 1..5, columns 1..2
M[2, 0, 3] = [0, 0, 0.2, 0.2, 0.2, 0, 0]        # explicit values along the last axis
B[:] = -1                                       # fill a range (a:b is half-open; ':' alone = the whole axis)
M[:, 1] *= 2                                    # scale (also += and -=)
```
Ranges: a:b is half-open, a:b:s takes every s-th index. Statements apply in order; a later statement overwrites an earlier
  one. Only the LAST edits block of your reply is read; a statement that cannot be read is skipped (the others still apply).
  Values: plain decimal numbers (|v| <= 1e6); a list must have exactly the shape of the indexed block.
\end{verbatim}
\end{minipage}
\caption{The complete prompt for SynthUterus, the only user message of each of its three calls (system prompt:
``Answer the user's request.''). Long lines are wrapped for display.}
\label{fig:prompt}
\end{figure*}

\paragraph{S2. An example answer.}
\Cref{fig:answer} shows a complete answer: Opus's first call for BreastMNIST (test AUC \exAnsAUC). The answer first describes the design in prose and then writes it as an edits block; only the
edits block is read. Its first line sets the output bias from the class prior, which the prompt does not state.

\begin{figure*}[p]
\centering
\begin{minipage}{\textwidth}\scriptsize
\input{tables/example_answer}
\end{minipage}
\caption{The complete answer of Opus's first BreastMNIST call, verbatim (its hidden reasoning is not part of the
answer). Long lines are wrapped for display, and Unicode minus signs, dashes and $\times$ are shown in ASCII.}
\label{fig:answer}
\end{figure*}
\fi
\end{document}